# First Clinical Experience in Urologic Surgery with a Novel Robotic Lightweight Laparoscope Holder

Jean-Alexandre Long (1), Jacques Tostain (2), Cecilia Lanchon (1), Sandrine Voros (3), Maud Medici (4), Jean-Luc Descotes (1), Jocelyne Troccaz (3),  Philippe Cinquin (3), Jean-Jacques Rambeaud (1), Alexandre Moreau-Gaudry (4)

1- Grenoble University hospital, Urology Department (France)

2- Saint-Etienne University Hospital (France)

3- TIMC-IMAG/GMCAO Lab (CNRS 5525) (France)

4- Clinical Investigation Center, Grenoble University Hospital, Grenoble (France)

Word count: 2691

Abstract: 240 words

**Address for correspondence**

Dr Jean-Alexandre LONG

Urology Department

Grenoble University Hospital

38 043 GRENOBLE Cedex 9

FRANCE

# Abstract

**Purpose**: To report the feasibility and the safety of a surgeon-controlled robotic endoscope holder in laparoscopic surgery.

**Materials and methods**: From March 2010 to September 2010, 20 patients were enrolled prospectively to undergo a laparoscopic surgery using an innovative robotic endoscope holder. Two surgeons performed 6 adrenalectomies, 4 sacrocolpopexies, 5 pyeloplasties, 4 radical prostatectomies and 1 radical nephrectomy. Demographic data, overall set-up time, operative time, number of assistants needed were reviewed. Surgeon's satisfaction regarding the ergonomics was assessed using a ten point scale. Postoperative clinical outcomes were reviewed at day 1 and 1 month postoperatively.

**Results:**

The per-protocol analysis was performed on 17 patients for whom the robot was effectively used for surgery. Median age was 63 years, 10 patients were female (59%). Median BMI was 26.8. Surgical procedures were completed with the robot in 12 cases (71 %). Median number of surgical assistant was 0. Overall set-up time with the robot was 19 min, operative time was 130 min) during which the robot was used 71% of the time. Mean hospital stay was 6.94 days ± 2.3. Median score regarding the easiness of use was 7. Median pain level was 1.5/10 at day 1 and 0 at 1 month postoperatively.

Open conversion was needed in 1 case (6 %) and 4 minor complications occurred in 2 patients (12%).

**Conclusion:**

This use of this novel robotic laparoscope holder is safe, feasible and it provides a good comfort to the surgeon.

# Introduction

The advantages of minimally invasive surgery are now well documented and laparoscopy is challenging for both surgeon and assistant. Manual control during prolonged cases can be exhausting either for the assistant or the surgeon who need a stable image. Among surgeons, urologists have early understood that robotic assistance could provide better comfort and surgical skills improvement. Despite the great advantages offered by the DaVinci® system (intuitive Surgical, Sunnyvale, US), this method remains expensive and cumbersome [1].

Our group developed a robotic lightweight endoscope holder that is now marketed by the company Endocontrol™ under the name ViKY® [2]. Previous cadaveric studies have shown the feasibility of the robot's use [3]. ViKY obtained CE marking in 2007 and FDA approval in 2008 [2]. The first use of the robot occurred in our institution on July 5, 2007. The first procedure was a bilateral pelvic lymph node dissection in the context of high grade prostate cancer previously to external beam radiotherapy.

The aim of this pilot study is now to assess the feasibility and the safety of this innovative medical device in different urological surgical indications. We present the results of the first clinical trial carried out with this VIKY robot prior to a bicenter randomized clinical study.

## Materials and methods:

### Study design

From March 2010 to September 2010, all the patients scheduled for a laparoscopic procedure in 2 institutions (Grenoble University Hospital and St Etienne University Hospital) were proposed to enrol the study. The clinical trial was approved by the French Ethical Committee CPP Sud-Est V. Exclusion criteria were: age < 18 years, pregnancy and inability to sign in the informed consent. Informed consent was obtained from 20 patients. All data were recorded in a case record form that was specifically designed for this study.

### Surgery description

Two laparoscopic surgeons (one in Grenoble and one in St Etienne) performed robotic surgeries for different clinical indications: adrenalectomy, acrocolpopexy, pyeloplasty, radical prostatectomy and radical nephrectomy. Nevertheless, prior to the start of this study, the surgeon had to perform three surgeries using the robot which gave him some experience in manipulating the robot. All the procedures used a transperitoneal approach.

### Robot description

The robot used in this trial consists of a compact motorized scope holder placed directly on the patients' abdomen **(figure 1, 2 and 3)**. Its architecture is based on a rotating circle. It is attached to the rail of the operating table using an articulated arm to improve the steadiness and the stability of the image. The endoscope manipulator is sufficiently small (110 mm in diameter) to be placed directly on the patient without interfering with other handheld instruments during minimally invasive surgery. It is 75 mm high and its weight is 625 g. The robot motors provide 3 degrees of freedom: 2 rotations allowing exploring the entire abdominal cavity, and translation allowing the endoscope to get closer to or further from the organs. It can be attached to any types of endoscopes and trocars. The robot is submersible and autoclavable. A console, which contains motor controllers and software, analyzes the surgeon's orders and translates them to commands for motors. It contains a touch panel screen for user interface. The system is controlled either by voice (Bluetooth microphone supervised by a single footswitch for security) or foot (6-function footswitch). The motors are back-drivable to allow a manual repositioning. Although the robot had been created for use in the dorsal supine position, it had the ability to extend the potential positions to include the lateral position.

**Data assessment**

Demographic data were recorded. Setup time including port placement and all the procedures until robot docking was reviewed as well as operative time, dismantling time and length of hospitalisation. Any robot technical problems or breakdown, manual completion or open conversion were recorded. After completion of each procedure, the surgeon's satisfaction was evaluated using a self 10 point scale to assess easiness of use, overall comfort, quality of vision and steadiness of image. Complications were separated as intraoperative or postoperative. The latter were classified according to the Clavien-Dindo classification [4]. Postoperative pain was evaluated using a 10 point scale the day after surgery and 1 moth after. We also used the painkiller records to evaluate the pain. Data were expressed as median and interquartile range or mean ± standard deviation in case of normal distribution. The descriptive statistical analysis was performed with GNU R software, version 2.13.

# Results

## Study population

Although 20 patients were enrolled in this clinical study, only 17 were involved in a robotic-assisted surgery : one patient has changed his mind before the surgery with a withdrawal of his consent (Grenoble) ;  for  another patient, a material malfunction of the robot was pre-operatively identified during the automatic functional check (hands-free headset problem) so that the robot could not be used during the surgery with a combination of the voice and foot, as wanted by the surgeon (Grenoble) ; lastly, the robot was not used in one case because of a difficult adhesiolysis requiring an human help (St Etienne). Per-protocol analysis is then performed on 17 patients for whom the surgery was effectively performed with ViKY.

## Patient's description

Demographic data are summarized in **table 1**. Median age was 63 years (IQR: 58-70). Ten patients were female (59 %). The median pre-operative BMI was 26.8 kg/m$^2$ (IQR: 25-28).

## Surgery description

Two laparoscopic surgeons performed 17 robotic surgeries: 2 sacrocolpopexies, 1 pyeloplasty, 2 radical prostatectomies and 1 radical nephrectomy were made in Grenoble. Six adrenalectomies, 1 sacrocolpopexy and 4 pyeloplasties were performed in St Etienne.

## Operative data

As summarized in **table 2**, the median operative time was 130 minutes (IQR: 110; 204) with a median overall setup time (including the robot set-up) at 19 minutes (IQR: 16; 25), i.e. 14.6% of the whole operative time. As illustrated **figure 4**, this setup time seems to be different according to the surgical indication. Especially for the radical prostatectomy which required a manual step for peritoneal incision and bladder detachment. Furthermore, as the same surgery is performed two times in a consecutive way, we observed a median time gain of 4.5 min, i.e. a relative gain time of 23%.The median assistant number was 0 per intervention (IQR: 0-1). Although the robot has fulfilled its role in 17 surgeries, 5 (29%) were not completed in their totality with the robot: a malfunction of the voice control occurred requiring to interrupt the robotic assisted surgery and four were interrupted because the surgical conditions required the help of a human assistant. After a robotic assisted pyeloplasty, one conversion to open surgery was performed to redo a pyeloureteral anastomosis. In 13 procedures (77%) the robot was voice controlled whereas it was pedal and voice controlled in 4 surgeries. Robots dismantle took a median time of 2 minutes (IQR: 1-4). The dismantle time seems to be linked to the surgical staff that uses the medical device (**Figure 5a and 5b** that illustrates different trends between Grenoble and St Etienne). One intraoperative complication occurred that was not due to the robot (pyeloureteral anastomosis leakage).

**Post operative outcomes**

Four grade II complications [1] occurred in 2 patients (12%): a patient had acute prostatitis associated to a haematoma after an adrenalectomy and another one had a bacteraemia after a urinary leakage. Patients were discharged from the hospital after a mean of 6.94 days $\pm$ 2.3. No skin damage was observed.

**Surgeon's satisfaction**

The scores of easiness of use was 7 (IQR: 4-9), global comfort was 7 (IQR: 5-8), quality of the vision was 9 (IQR: 7-9) and steadiness was evaluated to 10 (IQR: 8-10).

## Discussion

Among robots available for laparoscopic surgery, endoscope holders are designed to provide a steady, tremor-free image and a better visualization during the entire surgical procedure [5]. Surgeons themselves can direct their optical field, while the robot allows precise voice-activated, hand or foot control of the robotic camera holder.

ViKY robot had been designed by our group. This robot had been successfully validated through preclinical trials that showed the feasibility of the system in terms of workspace as well as compatibility of the system with an operating room environment on cadaver experiments [3,6]. The robot was then improved and upgraded on animal models.

This present study was designed to evaluate feasibility and safety of the use of this novel robotic endoscope holder for different urological surgical indications on human prior to a randomized control study whose inclusions began in September 2010.[7]

In this pilot study, we show that the robot is safe and user-friendly in human patients. Furthermore, the robot was easy and quick to set up and to dismantle in case of emergency. In this study, the learning curve was assessed by the dismantling time as the procedure is similar regardless to the intervention. The data show that in a team involved in the first steps of the robot, there was no improvement with the time suggesting a reproducible procedure whereas, this time quickly improved in a team not experienced with this robot. Consequently, we can assume that the main advantage of this system is its ease of use.

The advantage of the robot was the surgeon's complete autonomy over camera control. He didn't have to rely to the skill of an assistant. Although most of the procedures described require an active assistant during all steps, a couple of procedures such as nephrectomy and pyeloplasty could be performed by the surgeon all by himself. This is shown in our study where the assistant's median number was 0.

The only intraoperative complication reported was a urinary leak during a pyeloplasty requiring an open conversion. It was absolutely independent of the use of the device. Postoperative complications were rare. None of them could be connected to the use of the robot. We reported a bacteraemia after a urinary leakage, an acute prostatitis and a haematoma after an adrenalectomy that were all managed medically.

The reliability of the robot is a crucial issue. During this clinical trial, one robotic surgery was interrupted because of a material issue. Voice recognition was useful to control the endoscope's position. However, one robotic surgery had to be cancelled because of a failure of the system detected prior to the surgery, requiring a setting. This case was excluded in the per-protocol analysis. In a second case, the microphone broke down during surgery. Consequently, the robot was dismantled although the procedure could have been continued with the footswitch available. These shortcomings were corrected with the current version of the robot and a different positioning mechanism and actuators were installed to improve the reliability.

In 4 cases, the robotic procedure had to be interrupted indeed some limitations of motions in extreme positions exist with this robot

The robot can hamper the surgeon when wide motions are needed and when moving to an extreme upper position is required. Best surgeries for the endoscope holder are the one requiring few endoscope motions. However, this is the case with almost all the robots available including the DaVinci® system. It appears to us that in the urological field, the best indications seem to be pelvic surgery and adrenalectomy as the field of view is highly restricted. During a procedure, depending on the anatomical conditions, the range of motion can exceed the possibilities of the robot. As a result the robot needs to be replaced by a human assistant who is able to anticipate the surgeon's desired view without instruction especially when unexpected haemorrhage occurs. In our series, the robot had to be replaced in 4 cases.This high rate of robot's retrieval in our series can also partially be explained by the lack

5    of experience with the robot given the novelty of the device. Since this robot was designed by our group, one of the 2 surgeons was involved in the preclinical development of the device leading to an inherent bias concerning ergonomics evaluation. However one surgeon, located in an outside hospital, was totally novice in the use of the robot before inclusion in the study. This difference explains why no dismantle time learning curve was observed in the Grenoble

10   group although a time improvement was shown in the Saint-Etienne group. A potential drawback of the robot is the circular disk located at its base. This disk measures 10 cm. We showed that port placement had to be modified in 25% of the cases to avoid interferences between laparoscopic instruments and the robot. No consequence was reported due to the difference of port placement.

15   In comparison with existing robotic camera holders, the LER presents two advantages. First, its compactness compared to the other robots available (LapMan® (MedSys, Gembloux, Belgium) and Endoassist® (Armstrong Healthcare Ltd., High Wycombe, Buck, UK)) has a major influence on the acceptance in the operating room [12-14]. The robot AESOP that was the first endoscope holder created is no longer available. This robot has been the proof of the

20   concept that using an endoscope holder was feasible. Its diffusion has been restricted by Intuitive Surgical™ that purchased the company that sold this robot (computer Motion)[15]. This cumbersome robot had some limitations. Nevertheless its acceptation by many teams was excellent. These robotic endoscope holders cannot be compared to full robotic systems such as DaVinci®. Due to high definition, three dimensional optics and wristed instruments,

25   DaVinci assistance may be particularly well suited for tackling difficult surgeries in a minimally invasive manner. However their prohibitive cost is a limitation to their extensive use. The design of robotic endoscope holders aims to provide a low-cost, compact and lightweight system to help the surgeon during a standard laparoscopic procedure. Consequently, their objective is not to improve the dexterity of the surgeon but to provide to the surgeon a stable image and the capability to perform a solo-surgery in selected cases. Robotic endoscope holders are alternatives to static endoscope positionners such as Endofreeze® (Aesculap, Tuttlingen Germany) or Endoboy® (Geyser-Endobloc, Coudes, France) [5, 16]. These static positionners provide a stable image, but are inherently slower in use compared to robotic systems since the operator must release instruments and interrupt the procedure to make operating field adjustments.

As for robotic endoscope holders in general, the question is whether use of such a device is really necessary and if it is useful to provide to the surgeon the possibility to perform solo surgery without any human assistant [5]. Several studies comparing the differences between human and robotic control of the laparoscope have shown that the robotic system could be superior in terms of image steadiness [10]. However very few of these studies were prospective and randomized [10, 11]. We believe that each robot has its own indications. Concerning radical prostatectomy, that is the most common laparoscopic procedure in urology, a laparoscope holder can not compete with a full robotic mostly due to the technical advantages offered by the articulated robotic instruments of a DaVinci system. Such a laparoscope holder represents an available alternative to a human help or to allow the assistant to use his two hands to grab instruments and provide a real 4 hands surgery.

There were several limitations to the present study. This is a pilot study that can only assess the feasibility and the safety of the procedures using the robot. Further studies should investigate the clinical impact. Comparisons between standard laparoscopy and robotic camera assistance were not the aim of this pilot study. These first cases were included in this feasibility evaluation before a larger ongoing multi-institutional randomized and controlled study comparing different procedures performed with or without robotic camera assistance.

Furthermore, the surgeon's assessment of image stability and comfort is subjective even using a visual analog scale. The good results obtained in terms of image quality (8.5/10), image stadiness (10/10) contrast with a limited global comfort (6.5/10) and the high rate of manual completions (29%). These discrepancies could be explained by a bias induced by one surgeon's involvement in the development of this device.

**Conclusion**

In this pilot study, this novel robotic endoscope holder was evaluated for the first time in urologic surgery on human. VIKY use is feasible and safe. However, the high rate of manual completion and robot's dismantling needs to be evaluated on a further randomized controlled study to assess its real usefulness.

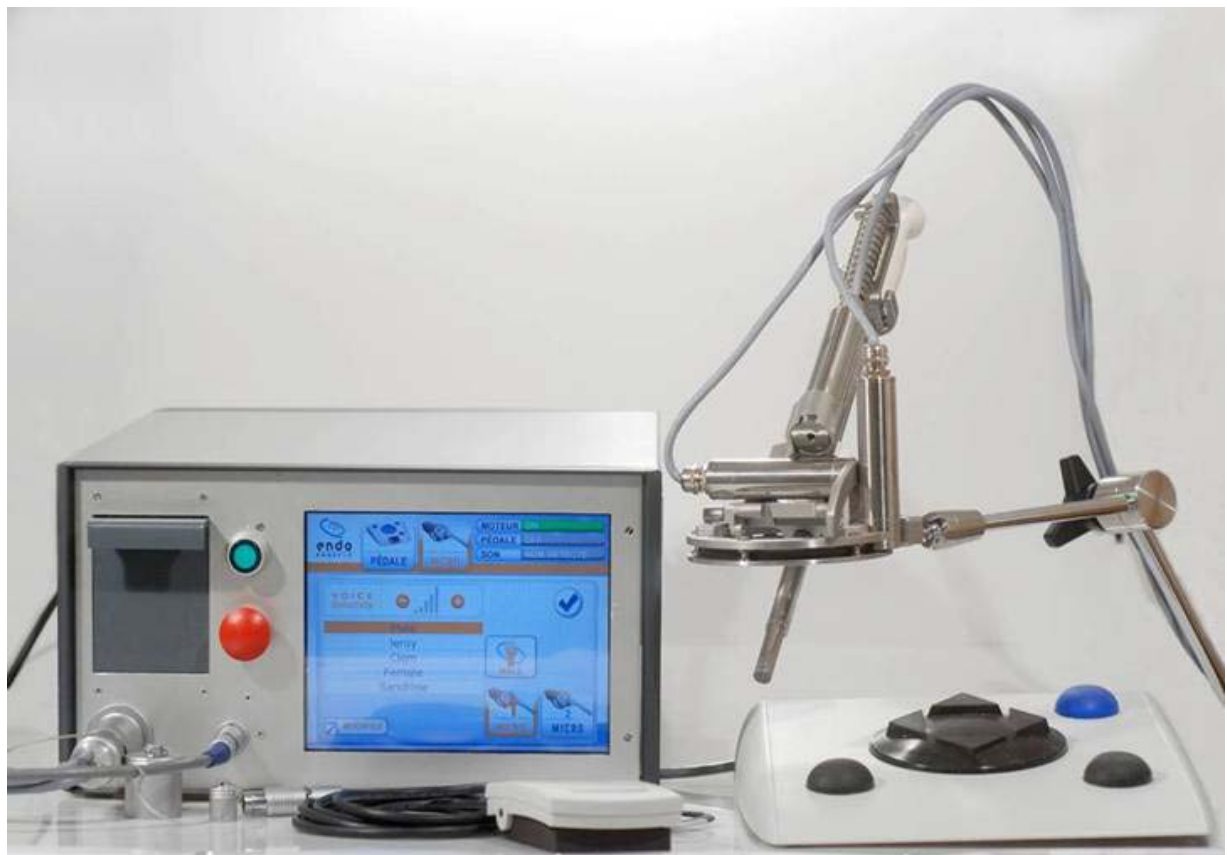

**Figure 1: The robotic system including the console, the robot and the pedal.**

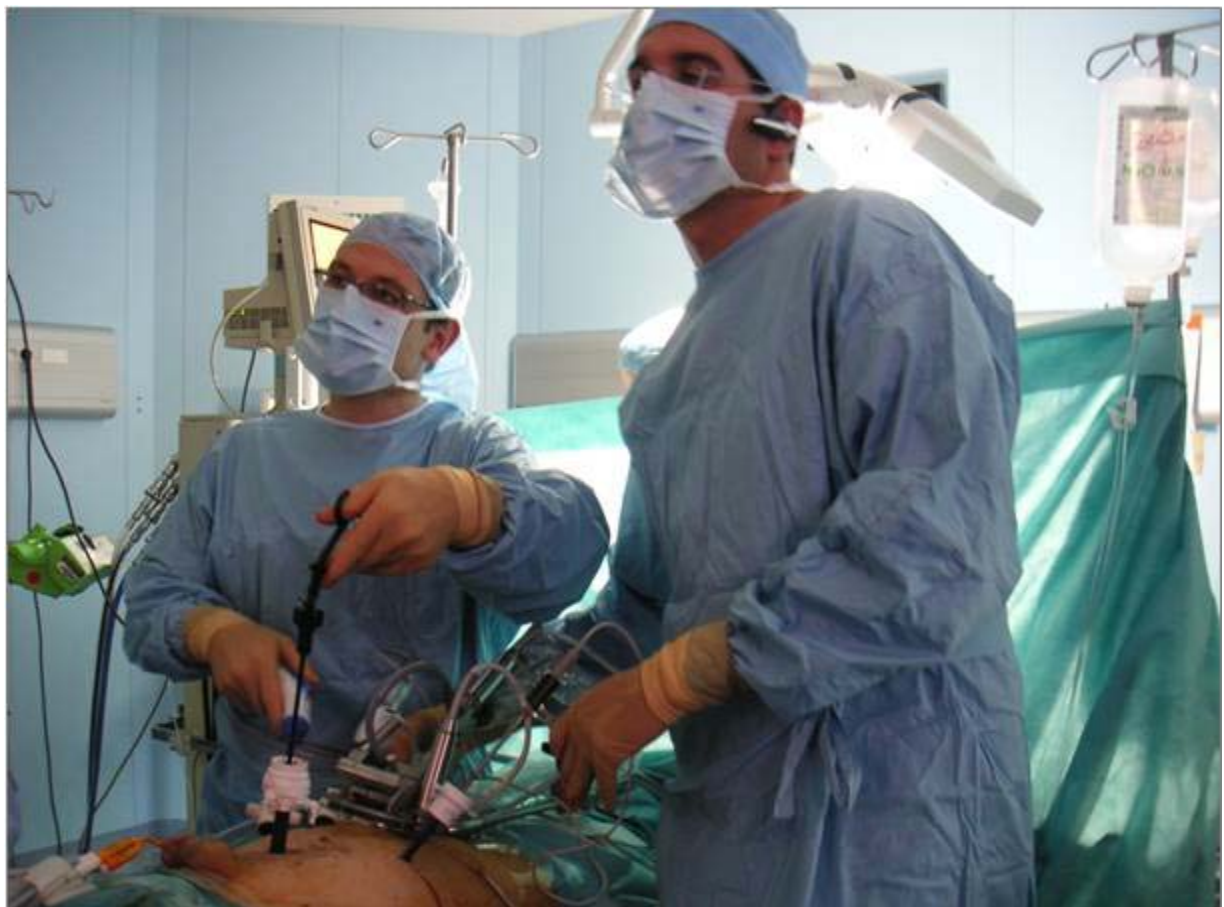

**Figure 2: The robotic system in use. Pay attention to the headset device allowing voice recognition.**

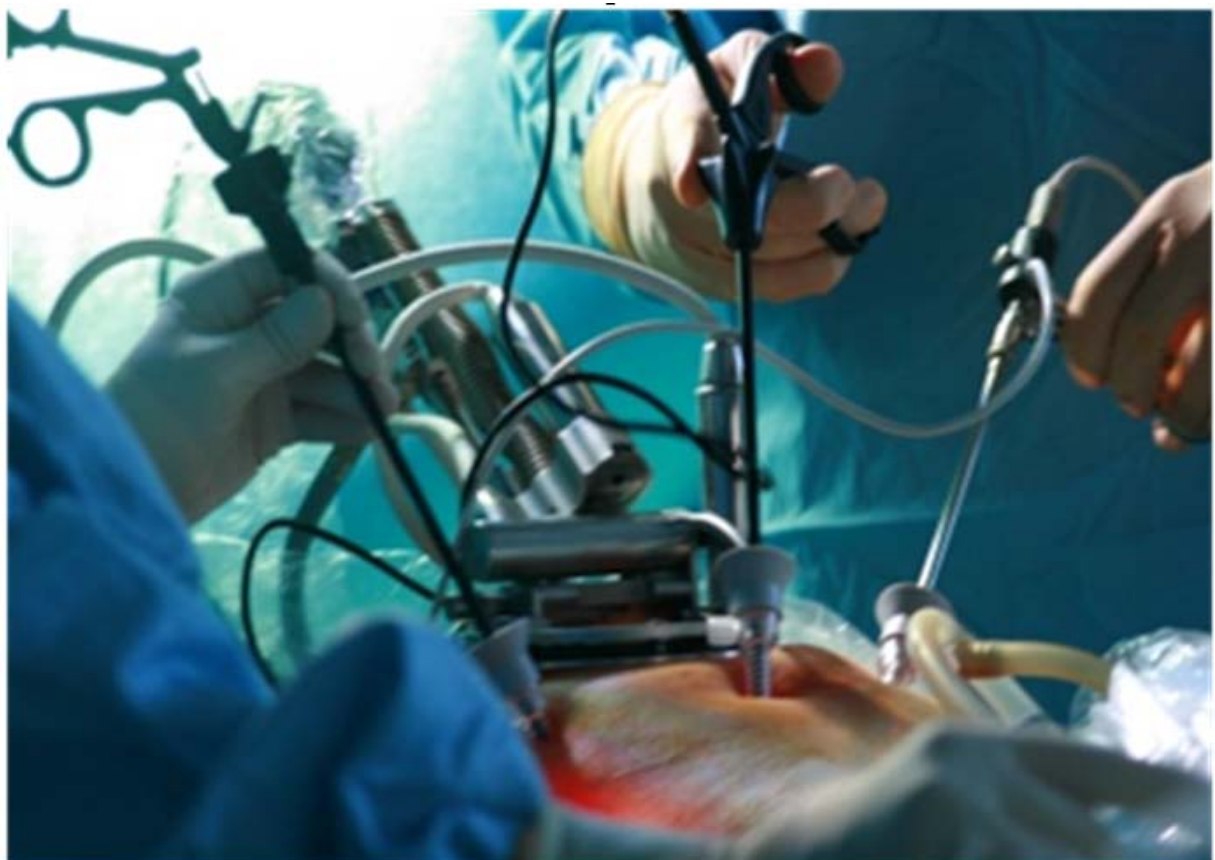

**Figure 3: The robot in the operative field**

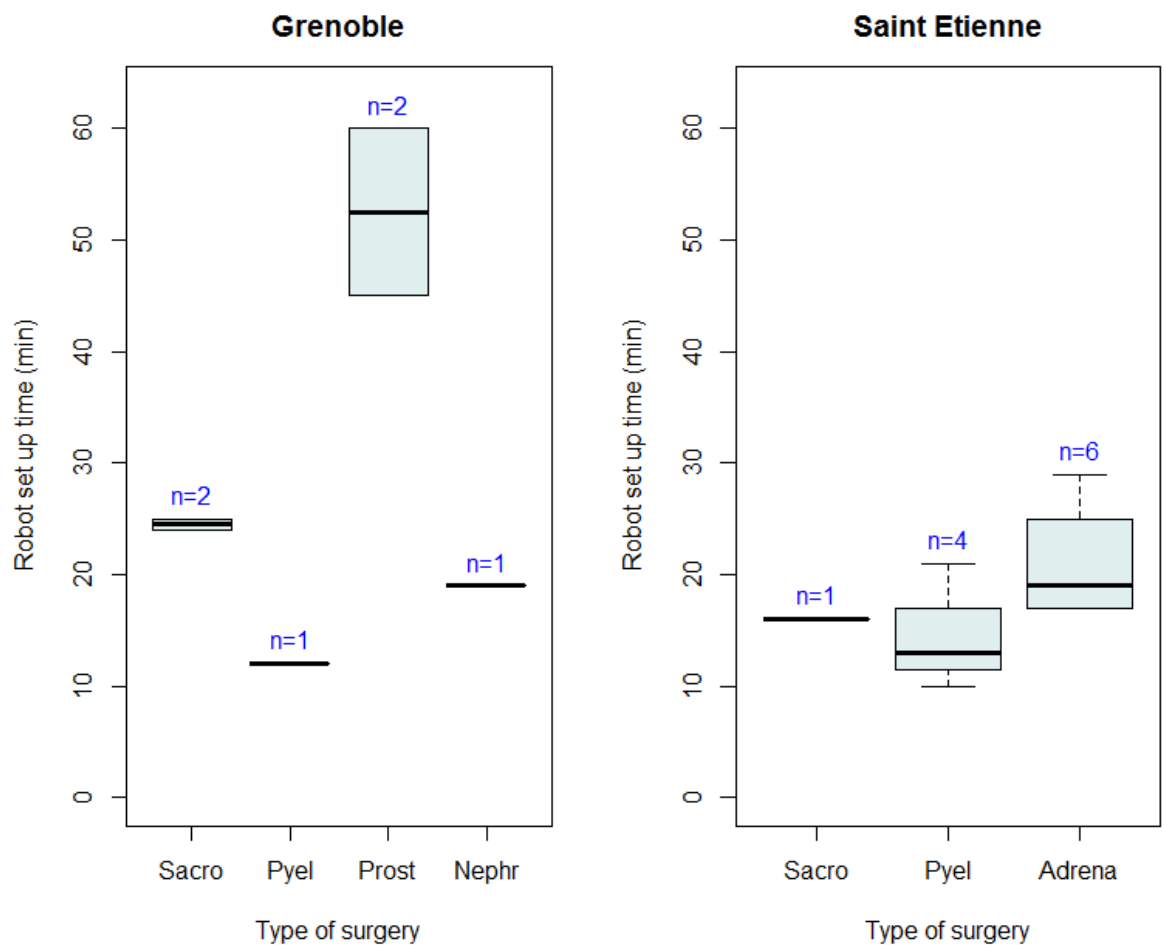

**Figure 4: Set up time (including port insertion and prior dissection to docking)**

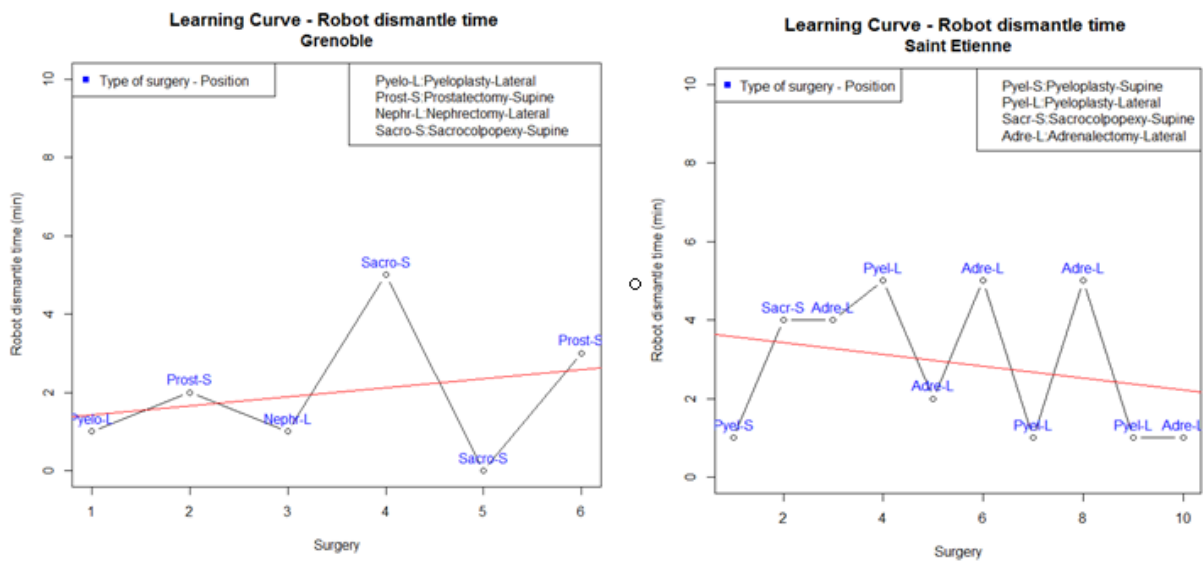

**Figure 5: Dismantling time: a- Grenoble, b- Saint-Etienne**

**Table 1: Demographic data and operative settings**

| | |
|---|---|
| **Median age (IQR)** | 63 (58-70) |
| **Female gender** | 10 (59%) |
| **Median BMI (IQR)** | 26.8 (25-28) |
| **ASA score** | |
| **1** | 4 (23%) |
| **2** | 8(47%) |
| **3** | 5 (29%) |
| **Surgery** | |
| **Sacrocolpopexy** | 3 (18%) |
| **Pyeloplasty** | 5 (29%) |
| **Radical Nephrectomy** | 1 (6%) |
| **Adrenalectomy** | 6 (35%) |
| **Radical Prostatectomy** | 2 (12 %) |
| **Position** | |
| **Supine** | 6 (35%) |
| **Lateral** | 11 (65%) |
| **Voice recognition use** | 13 (77%) |

**Table 2: Operative data and postoperative outcomes**

| | |
|---|---|
| **Median assistant number** | 0 (0-1) |
| **Median setup time** | 19 (16-25) |
| **Operative time (min)** | 130 (110-204) |
| **Median dismantling time (min)** | 2 (1-4) |
| **Mean length of stay (days)** | 6,94 ± 2,3 |
| **Median port number** | 4 (3-4) |
| **Robotic successful completion** | 12 (71%) |
| **Intraoperative complications** | 1 (6%) |
| **Postoperative complications** | 4 (23%) |
| **Pain Day 1 (10 point analog scale)** | 1,5 (0.8-3) |
| **Painkillers Day 1** | |
| **Level 1** | 3 (18 %) |
| **Level 2** | 4 (23%) |
| **Level 3** | 7 (41 %) |
| **Pain Month 1 (10 point analog scale)** | 0 (0-1.25) |
| **Surgeons satisfaction (10 point analog scale)** | |
| **Easiness of use** | 7 (6-9) |
| **Global comfort** | 7 (5-8) |
| **Image quality** | 9(7-9) |
| **Image steadiness** | 10 (8-10) |

Data are expressed as median (IQR) or mean ± SD or frequency (%)